\documentclass{article}

\usepackage{microtype}
\usepackage{graphicx}
\usepackage{subfigure}
\usepackage{booktabs}
\usepackage{hyperref}
\usepackage[accepted]{icml2024}

\usepackage{amsmath}
\usepackage{amssymb}
\usepackage{mathtools}
\usepackage{amsthm}
\usepackage[capitalize,noabbrev]{cleveref}

\usepackage[utf8]{inputenc}
\usepackage[T1]{fontenc}
\usepackage{amsfonts}
\usepackage{nicefrac}
\usepackage{xcolor}
\usepackage{verbatim}
\usepackage{xspace}
\usepackage{pifont}
\usepackage{enumitem}
\usepackage{url}

\newcommand{\cmark}{\ding{51}}
\newcommand{\xmark}{\ding{55}}

\begin{document}

\twocolumn[
\icmltitle{Re-evaluating Retrosynthesis Algorithms with Syntheseus}

\icmlsetsymbol{equal}{*}

\begin{icmlauthorlist}
\icmlauthor{Krzysztof Maziarz}{msr}
\icmlauthor{Austin Tripp}{cambridge}
\icmlauthor{Guoqing Liu}{msr}
\icmlauthor{Megan Stanley}{msr}\\
\icmlauthor{Shufang Xie}{msr}
\icmlauthor{Piotr Gaiński}{jagiellonian}
\icmlauthor{Philipp Seidl}{jku}
\icmlauthor{Marwin Segler}{msr}
\end{icmlauthorlist}

\icmlaffiliation{msr}{Microsoft Research AI for Science}
\icmlaffiliation{cambridge}{University of Cambridge}
\icmlaffiliation{jagiellonian}{Jagiellonian University}
\icmlaffiliation{jku}{Johannes Kepler University Linz}

\icmlcorrespondingauthor{}{}

\icmlkeywords{Machine Learning, AI for Science, Retrosynthesis, Synthesis, Planning}

\vskip 0.3in
]

\printAffiliationsAndNotice{}

\begin{abstract}
Automated Synthesis Planning has recently re-emerged as a research area at the intersection of chemistry and machine learning.
Despite the appearance of steady progress, we argue that imperfect benchmarks and inconsistent comparisons 
mask systematic shortcomings of existing techniques, and unnecessarily hamper progress.
To remedy this, we present a synthesis planning library with an extensive benchmarking framework, called \textsc{syntheseus}, which promotes best practice by default,
enabling consistent meaningful evaluation of single-step models and multi-step planning algorithms.
We demonstrate the capabilities of \textsc{syntheseus} by re-evaluating several previous retrosynthesis algorithms,
and find that the ranking of state-of-the-art models changes in controlled evaluation experiments.
We end with guidance for future works in this area, and call the community to engage in the discussion on how to improve benchmarks for synthesis planning.
\end{abstract}

\section{Introduction}

Since the advent of computers, chemists have envisioned automating synthesis planning~\citep{vleduts1963concerning, corey1969computer}. Automated Synthesis Planning has two main uses: First, it can help chemists find better retrosynthetic routes faster, by providing route suggestions for synthetic problems at hand. Second, it can be used to predict synthesizability in computational discovery workflows for large numbers of molecules. 
This use has become increasingly relevant via the renewed interest in algorithmic de novo molecular design, largely driven by generative models for molecules over the past seven years~\citep{segler2017generating,gomez2018automatic,meyers2021novo,maziarz2021learning}. 
Although de novo design can help to discover compounds with desired property profiles more efficiently,
most existing methods do not explicitly account for synthesizability, and therefore
often output molecules that are hard to make in a wet-lab. This can limit their use in molecular discovery, where reliable synthesis is paramount~\citep{klebe2009wirkstoffdesign}.
Fast and reliable algorithms
that check for synthesizability by explicitly planning synthesis routes for an input molecule would allow to remedy this issue~\citep{stanley2023fake}.

Computer-Assisted Synthesis Planning (CASP) has a long tradition in chemoinformatics research, going back to the 1960ies with the visionary works of~\citet{vleduts1963concerning} and~\citet{corey1969computer}. Most early proposed approaches were based on rule-based expert systems, with rules either compiled by human experts or automatically data-mined~\citep{todd2005computer, cook2012computer, szymkuc2016computer}.
Most recently, conceptual progress in synthesis planning has been achieved through the use of deep neural networks, which learn to predict feasible and rank the most viable disconnections, and guide the search algorithms into promising directions~\citep{segler2018planning,  strieth2020machine, schwaller2022machine, tu2023predictive}.
Surprisingly, most published work on synthesis planning prior to deep learning did not feature quantitative metrics to systematically improve models, but rather provided a limited number of qualitative test cases as proof.

Synthesis planning, also known as retrosynthesis, usually works by recursively decomposing a target molecule into increasingly simpler molecules using formally reversed chemical reactions, until a set of purchasable or known building block molecules is found. Starting with the building blocks, the reactions in the forward direction provide a recipe of how to synthesize the target. 
Computational tools for synthesis planning therefore generally feature three components: (1) a single-step retrosynthesis model or component providing possible reactions that could give rise to a given molecule; (2) a search algorithm that strings together single-step reactions into a multi-step route, which can be ranked according to a given criteria such as cost or number of steps; and (3) a set of allowed building blocks and starting materials from which the molecules are constructed. An alternative to searching backwards is to perform goal-directed forward synthesis prediction~\citep{bradshaw2020barking, gao2021amortized}, an emerging approach, which has the potential to also be used in bi-directional search~\citep{ihlenfeldt1996computer}.
Commonly, single-step retrosynthesis models and multi-step planning algorithms are studied independently.

\begin{figure*}[h!]
\vskip-0.25cm
\centering
\includegraphics[width=0.8\textwidth]{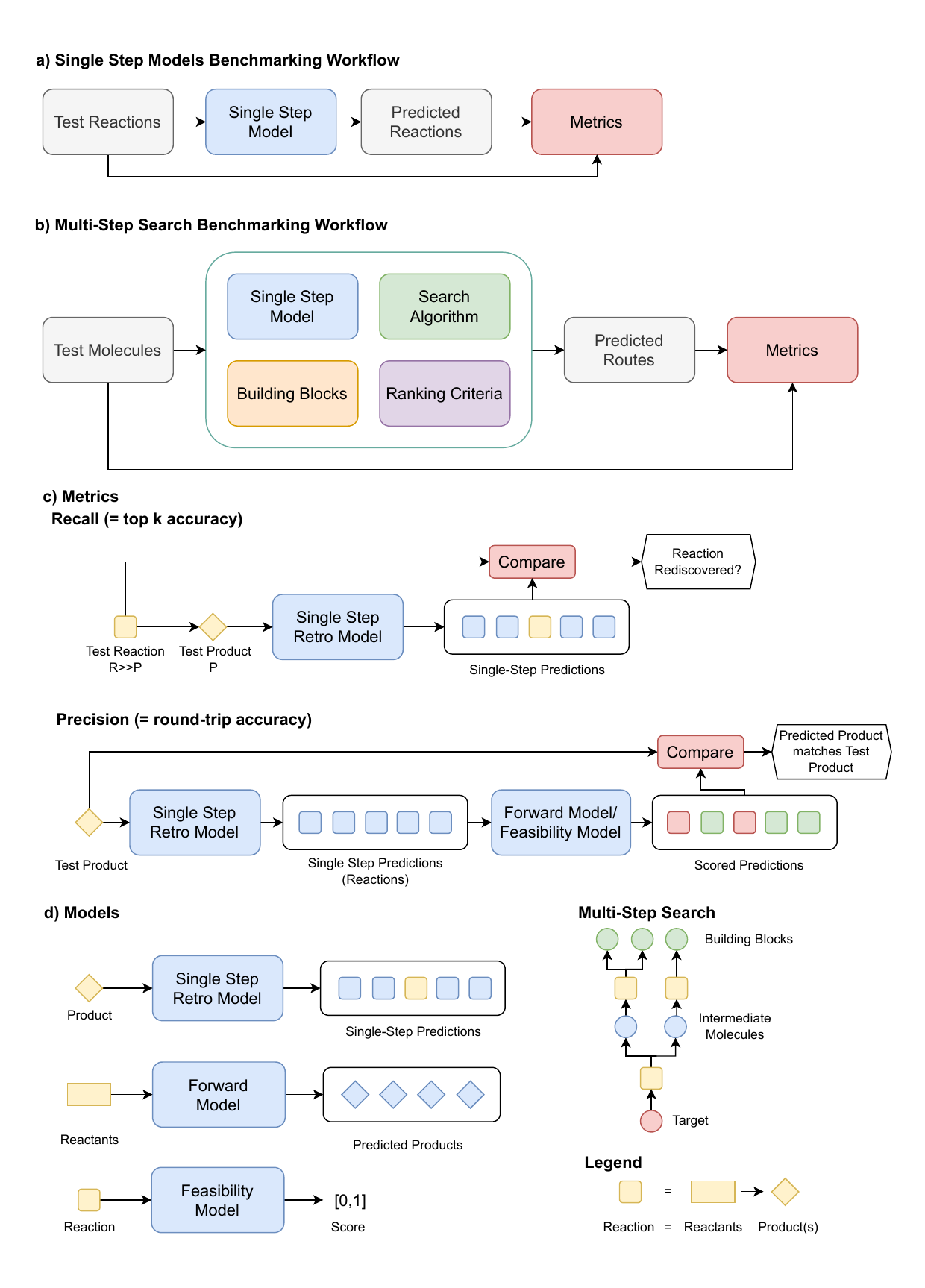}
\vskip-0.1cm
\caption{Benchmarking workflows and metrics studied in this work.}
\label{fig:overview}
\end{figure*}

\looseness=-1
In \emph{single-step retrosynthesis}, models are given a molecule and output reactions which produce that molecule in one step~\citep{segler2017neural, coley2017retrosim, liu2017s2s, dai2019retrosynthesis, tetko2020state, yan2022retrocomposer, wang2023retrosynthesis,igashovretrobridge}.
Recent work in this area has generally focused on training neural networks to predict reactions extracted from the scientific literature or patents~\citep{lowe2012extraction, zhong2023recent}.
In \emph{multi-step planning}, given a target molecule,
a set of purchasable molecules, and a single-step retrosynthesis model, the goal is to produce complete synthesis routes.
This is challenging as the search space is extremely large compared to the number of solutions.
Recent work in this area has used Monte Carlo Tree Search (MCTS), Reinforcement Learning (RL), or heuristic-guided search algorithms, to selectively explore a tree of possible reactions from the starting molecule~\citep{segler2018planning, coley2019robotic, schwaller2020predicting, chen20retrostar, xie2022retrograph, tripp2022reevaluating, liu2023retrosynthetic,tripp2023retro}.

\looseness=-1
In this work, we take a closer look at the commonly used metrics for single and multi-step retrosynthesis.
First, it is not clear how metrics used when benchmarking single-step and multi-step in isolation should be interpreted in the context of an end-to-end retrosynthesis pipeline.
Second, 
model comparison and metrics use in prior work has been inconsistent. 
The goal of this paper is to specify best practices for evaluating retrosynthesis algorithms, which we codified in a python package called \textsc{syntheseus}, allowing researchers to evaluate their approaches in a consistent way. A depiction of the different models, metrics and benchmarking workflows is given in Figure~\ref{fig:overview}. The capabilities of the package are demonstrated via a re-evaluation and analysis of prior work. 
The paper is organized as follows:

Sections~\ref{sec:related work} and \ref{sec:discussion} examine how retrosynthesis was evaluated in previous works, and point out shortcomings in this practice.
We find several previously reported results to be understated, overstated, or otherwise not comparable to each other. We then outline best practices for the field.
In Section~\ref{sec:syntheseus} we present our python package \textsc{syntheseus}.
It supports consistent evaluation of single-step and multi-step retrosynthesis algorithms, with best practice enforced by default.
In Section~\ref{sec:experiments} we use \textsc{syntheseus} to re-evaluate existing single-step and multi-step methods across many settings in an attempt to ``set the record straight''. 
We also highlight remaining gaps where existing metrics are not sufficient. 
We envision that our results will serve as a starting point for future research, which we hope \textsc{syntheseus} to accelerate.
In Section~\ref{sec:summary} we provide a roadmap of how \textsc{syntheseus} can contribute to longer-term improvements in retrosynthesis methods and their evaluation.

\section{Prior Work on Benchmarking}\label{sec:related work}

\looseness=-1
Several works have published benchmarks for retrosynthesis.
For single-step, the USPTO-50K dataset~\citep{schneider2016s} is the most popular,
while for multi-step many papers report results on the 190 hard molecules from Retro*~\citep{chen20retrostar}.
However, these benchmarks do not have a standardized evaluation pipeline,
leading to inconsistent re-implementations by different authors often subject to pitfalls discussed in Section~\ref{sec:discussion} (particularly S\ref{ssr:post-processing}, S\ref{ssr:data-leakage} and M\ref{msr:single-step-model}, M\ref{msr:cache}).
\textsc{Syntheseus} allows these benchmarks to be
run in a consistent and comparable way.
Nonetheless, these benchmarks are far from perfect:
their standard metrics include recall (S\ref{ssr:precision-recall})
and success rate (M\ref{msr:success-rate})
but do not include inference time (S\ref{ssr:inference-time}) or diversity (M\ref{msr:diversity}).
In a major step forward, \citet{genheden2022paroutes} proposed the PaRoutes benchmark for multi-step search,
which does include an assessment of diversity (M\ref{msr:diversity}) and a standardized evaluation script.
Unfortunately, it measures diversity with the output of a clustering algorithm:
a metric which is non-monotonic.
This makes it possible for an algorithm to find strictly more routes than another algorithm
yet be rated as less diverse.
In contrast, the diversity metric employed in this work is monotonic,
meaning that finding additional routes will never cause diversity to decrease.

\looseness=-1
More broadly, some works have highlighted the deficiencies with retrosynthesis evaluation.
\citet{thakkar2020datasets}, \citet{schwaller2020predicting} and \citet{zhong2023recent} point out how separate evaluation of single-step and multi-step
may not lead to effective CASP programs and mention the limitations of recall (top-$k$ accuracy) (S\ref{ssr:precision-recall}).
\citet{segler2018planning} noted the inherent shortcomings of \textit{in-silico} evaluation
and benchmarked their algorithm with an A/B or Turing test by expert chemists.
However, evaluation through human feedback is cumbersome, making such examples rare in ML venues.
\citet{hassen2022mind} correctly noted that the performance of multi-step search algorithms will depend on the single-step model
and performed an evaluation of many single-step models combined with popular multi-step search algorithms;
this analysis was later extended in~\citet{torren2023models} to large proprietary training datasets.
However, these works quantitatively compare the results across different single-step models using success rate, which we argue is not sufficient (M\ref{msr:success-rate}).

\looseness=-1
Finally, it is worth mentioning several popular software packages for retrosynthesis. ASKCOS~\citep{coley2019robotic} and
AiZynthFinder~\citep{genheden2020aizynthfinder} are software packages for multi-step search with
a simple interface and interactive visualizations.
However, they are primarily designed to support MCTS with template-based models, even though lately extensions have been proposed.
In contrast, \textsc{syntheseus} is designed from the start in a model-agnostic and algorithm-agnostic way,
and is easy to extend to arbitrary models and algorithms.
IBM RXN, Molecule.One and Chematica~\citep{klucznik2018efficient} are popular software tools for retrosynthesis,
but unlike \textsc{syntheseus} cannot be used for benchmarking as they are closed-source.
This calls for an open-source synthesis planning platform with fully-featured benchmarking capabilities, which we hope to provide with \textsc{syntheseus}.

\section{Pitfalls and Best Practice for Retrosynthesis Evaluation}
\label{sec:discussion}

Evaluation in retrosynthesis is largely constrained by two realities.
First, actually performing synthesis in the lab is costly, time-consuming,
and requires synthetic chemistry expertise; it is therefore infeasible for most researchers
who work on algorithm development, and should not be a requirement,
even though experimental validation is clearly important.
Second, because of the division into single-step and multi-step,
most works seek to evaluate one part of the retrosynthesis pipeline in isolation
rather than holistically,
while the key to real-world adoption lies in end-to-end performance.
Keeping this in mind, in this section we survey the merits and shortcomings of existing evaluation practices for single-step models and multi-step search algorithms.

\subsection{Single-Step Models}

\newcounter{single-step-recs}

Single-step retrosynthesis models have several functions in CASP programs: (1) defining which reactions are \emph{feasible} ways of obtaining a given molecule in one step, effectively defining the search environment; and (2) \emph{ranking} or otherwise expressing preference over these reactions, effectively acting as a policy or heuristic to guide the search.
Most single-step retrosynthesis models output a list of reactions, and are trained
using supervised learning to output reactions which were used in real synthesis routes
and furthermore to rank these reactions highly. 
The most common evaluation strategy is to compute the top-$k$ accuracy on a held-out test set,
i.e.\@ the fraction of molecules where the reaction which occurred in the dataset is ranked in the first $k$ outputs~\citep{segler2017neural, liu2017s2s, coley2017retrosim}.
This parallels evaluation commonly used in computer vision~\citep{deng2009imagenet, krizhevsky2017imagenet}.
In this section, we explain how this evaluation metric does not fully measure the utility of single-step models in CASP programs,
and how subtle differences in evaluation have distorted the numbers reported in prior works.
We suggest best practice for each of these points.

\refstepcounter{single-step-recs}\label{ssr:precision-recall}
\subsubsection*{Pitfall S1: Measuring only recall without considering precision}

By measuring how often reactions from the dataset occur in the model outputs,
top-$k$ accuracy essentially tests the model's ability to \emph{recall} the dataset. In particular, when using larger $k$, it is a useful, intuitive metric to assess whether a model can recover the full diversity of the dataset, and not just cover the most commonly used reactions. 
However, unless $k=1$ and the top-1 accuracy is nearly 100\%,
a multi-step search algorithm using the given single-step model will almost certainly
use reactions not contained in the dataset for planning.
If these reactions have low quality or feasibility then routes using them
will not be useful. On the other hand, in many cases there are several possible ways to make a particular molecule, while existing datasets are sparse, and usually only one or few possible syntheses have been reported for most molecules. 
Therefore, as previously argued by \citet{schwaller2020predicting}, top-$k$ \emph{precision} of a single-step model 
(what fraction of the top $k$ reactions are feasible)
is arguably equally or more important than recall for multi-step search.
Unfortunately, without an expert chemist or a wet-lab, precision is hard to measure.
Nonetheless, this suggests that models with a higher top-$k$ accuracy are not necessarily more useful
in CASP programs.\newline
\textbf{Best practice:} Besides recall, authors should strive to evaluate the precision of their models,
at the very least through a visual check of several examples.
Some prior works use round-trip accuracy using a forward reaction model (also referred to as back-translation) to measure feasibility of reactions that are not necessarily ground-truth~\citep{schwaller2019evaluation, chen2021deep}.
However, we note the inconsistent use of the term ``round-trip accuracy'' in prior work: \citet{chen2021deep} compute it in a top-$k$ fashion where \emph{at least one} of the top-$k$ results has to round-trip in order for the prediction to count as successful,
which does \emph{not} measure precision;
in \citet{schwaller2020predicting} this metric is called \emph{coverage}.
Round-trip accuracy also relies on a fixed forward model, which is usually only trained on real reactions (i.e. is given sets of reactants that actually react as input) without the presence of negative data;
it is unclear whether such a model can be used to evaluate reaction feasibility more broadly.
An alternative could be explicitly trained feasibility models, such as the binary filter models used by Segler and Coley~\citep{segler2018planning, coley2017prediction, coley2019robotic, gainski2024retrogfn}. 
In summary, while we encourage the use of round-trip accuracy using a forward model as a proxy,
we think this needs more attention and further developments from the community before reaching its full potential as a metric (see also M\ref{msr:success-rate} and M\ref{msr:chemist-eval}).

\refstepcounter{single-step-recs}\label{ssr:post-processing}
\subsubsection*{Best practice S2: use consistent and realistic post-processing}
Most prior works perform some amount of post-processing of model outputs
when measuring accuracy.
Unfortunately, this has not been done consistently by previous papers,
distorting comparisons between methods.
In general, the evaluation post-processing should match the post-processing that would be performed
if the model was used in a CASP program.
We identify several instances of this below and suggest best practice.
\begin{itemize}[topsep=0.1cm,itemsep=0.05cm,leftmargin=0.75cm]
    \item \textbf{Invalid outputs:} some models
        can output invalid molecules (e.g.\@ syntactically invalid SMILES strings)~\citep{irwin2022chemformer}.
        When computing top-$k$ accuracy,
        some prior works include invalid molecules in the top-$k$,
        whereas other works filter them out and consider the top-$k$ \emph{valid} molecules.
        As molecule validity is generally easy to check with chemoinformatics toolkits,
        a well-engineered CASP program would discard invalid molecules
        instead of considering them during search.
        Therefore, we believe best practice should be to only consider valid molecules when computing accuracy.
    \item \textbf{Duplicate outputs:}
        some models can produce the same result (set of reactants) multiple times.
        Clearly, a well-engineered CASP program would remove duplicate reactions, because they are redundant for search.
        However, this has not been done consistently in prior work.
        For example, we found that the published top-5 accuracy of GLN~\citep{dai2019retrosynthesis} on USPTO-50K can be increased by as much as 5.8\% with simple deduplication.
        Therefore, we think best practice is to measure accuracy \emph{after} deduplicating the outputs.
    \item \textbf{Stereochemistry:}
        Correct handling of stereochemical information in CASP is far from trivial, given the different possible ways in which stereocontrol can be achieved, e.g. via chiral pool synthesis, stereoselective reagents or catalysts, or resolution techniques. Nevertheless, as a proxy, many prior works require an exact match of stereochemistry
        in order for a prediction to count as correct. 
        Unfortunately, in the chemical literature and popular datasets like USPTO~\citep{schneider2016s}
        stereochemistry is often unlabelled or mislabelled,
        which motivated the authors of LocalRetro~\citep{chen2021deep} to measure a relaxed notion of accuracy where
        a prediction can be deemed correct even if its stereochemistry is different to the dataset.\footnote{In LocalRetro a prediction is considered correct if its set of stereoisomers is either a subset or a superset of the set of stereoisomers of the ground-truth answer; see \texttt{isomer\_match} in \href{https://github.com/kaist-amsg/LocalRetro}{github.com/kaist-amsg/LocalRetro}.}
        However, this practice was not applied to baselines LocalRetro was compared to, and subsequent authors copied the result from~\citet{chen2021deep} unaware that it uses a different definition of success. In our re-evaluation we found that using a relaxed comparison significantly boosted the reported accuracy of LocalRetro on USPTO-50K (e.g. +1.3\% top-1 and +2.6\% top-50); same is true for RetroKNN~\citep{xie2023retroknn} which built upon LocalRetro and re-used their evaluation code.
        While some datasets like USPTO-50K indeed contain chirality errors,
        in real-world scenarios CASP programs
        should not discard it;
        we therefore believe that best practice is to report the standard exact match, with potentially additional reporting of results with stereochemistry removed to provide further valuable insight.
\end{itemize}

\refstepcounter{single-step-recs}\label{ssr:inference-time}
\subsubsection*{Best practice S3: report inference time}
In contemporary ML works, it is common to give little attention to inference time, and focus on increasing the quantitative model performance. However, in retrosynthesis prediction, the purpose of a single-step model is to act as an environment during multi-step search. In practice, having a drastically faster single-step model can translate to doing a much more extensive search, thus single-step model speed is directly tied to quantitative performance downstream. Due to that, we believe future research should give more attention to accurately reporting inference speed, reasoning in terms of a speed-accuracy Pareto front rather than accuracy alone.
At the very least, we believe best practice is to report inference time in addition to accuracy, and also the hardware used to run the experiments.

\refstepcounter{single-step-recs}\label{ssr:unk-rxn-type}
\subsubsection*{Best practice S4: focus on prediction with unknown reaction type}
Most single-step works using USPTO report two sets of metrics: one for when the reaction type is not known, and another one for when the reaction type is given as auxiliary input; a practice started by~\citet{liu2017s2s}. The rationale for the latter usually involves an interactive setting where a chemist may prefer one reaction type over another. In the context of multi-step search this information would not be available, and it is unlikely that a given reaction type is universally preferred across the entire search tree.
In any case, none of the popular multi-step search algorithms add reactions conditioned on a particular reaction type, so this ``conditional reaction prediction'' would not be used by existing approaches.
Thus, our recommendation is for researchers to focus on the ``reaction type unknown'' setting, as this is the one most directly applicable to multi-step search.

\refstepcounter{single-step-recs}\label{ssr:data-leakage}
\subsubsection*{Best practice S5: avoid leakage through atom-atom mappings}
Most reaction datasets are annotated with atom-atom mappings (AAM), which map atoms from the reactant molecules to the corresponding atoms in the products; this is important for example to assign the reactive parts of the molecule or classify reaction mechanisms. To create an AAM, one needs to know both reactants and products of a reaction. However, in retrosynthesis prediction only the product is known a priori without any mapping, therefore such mapping is a potential source of data leakage. 
Indeed, some results on USPTO-50K were later found to be flawed for this reason due to unexpected behaviour of rdkit canonicalization, which is influenced by the presence or absence of atom mapping in test molecules, and thus from the AAM number the reacting atoms could easily be guessed~\citep{yan2020retroxpert,gao2022semiretro}.
While this problem is known to many practitioners, we mention it for completeness.
To avoid this pitfall, the input molecules should be provided to the model with the atom mapping information removed, and they have to be re-canonicalized \emph{after} said removal.

\subsubsection*{Pitfall S6: Dataset selection}
\label{ssr:data-set-size}
Researchers should choose their datasets according to the hypotheses they want to test.  
For production use, it is most common to train on the largest available (in-house) dataset, well-curated according to the needs of the respective organization. 
Unlike in other domains, such as images or language, large unlabeled reaction datasets are not yet publicly available. 
Therefore, for benchmarking, it is most common to train retrosynthesis models from scratch, and care must be taken to not make unrealistic assumptions on the selection of pre-training datasets to avoid data leakage. 
While USPTO-50K has been widely used as the "MNIST" for synthesis, and can be used as a minimum viable test, it is recommended to train and test on larger datasets whenever they are available. Additional care has to be taken when aggregating prior results from papers, as several different versions of datasets with the same name exist (e.g. for USPTO-FULL). While the community has started to recognize the need for better curated and standardized open datasets, more work in this area is needed~\citep{kearnes2021open,genheden2022paroutes,wigh2023orderly}.

\subsubsection*{Pitfall S7 / M0: Implementation matters}
Both single step models and search algorithms can be non-trivial to implement and tune. This applies particularly to more advanced template- or graph edit-based models, for which in contrast to Transformer models usually no well-optimized, general purpose codebases can be adapted. This can lead to authors reporting varying performance for seemingly the same models and algorithms~\citep{chen20retrostar, genheden2022paroutes}, as well as the perception of performance differences that diminish with increased engineering investment. 
Similar trends have been observed in the reinforcement learning literature~\citep{henderson2018deep}.

\subsection{Multi-Step Search}
\label{sec:discussion-multi-step}

\newcounter{multi-step-recs}

The role of a multi-step search algorithm is to use a single-step model,
a set of purchasable molecules,
and optionally some heuristic functions, in order to find synthesis routes.
Most prior works evaluate multi-step search algorithms by reporting the fraction
of test molecules solved in a given time window, where time is often measured
with the number of calls to the reaction model.
As in the previous section, here we explain the pitfalls and abuses of this metric and suggest best practice going forward.

\refstepcounter{multi-step-recs}\label{msr:single-step-model}
\subsubsection*{Pitfall M1: changing the single-step model without control}
Many algorithms use single-step reaction models not only to define the search environment,
but also use the rankings or probabilities from a single-step model as a policy, cost function, or to otherwise guide the search~\citep{segler2018planning, akihiro2019dfpns, chen20retrostar}.
Naturally this has led some works to modify the single-step model in order to improve search performance~\citep{kim2021self, yu2022grasp}.
These modifications not only change the relative rankings, but also the set of produced reactions.
We see two pitfalls with the way this has been used in practice.
First, unless the single-step model is separately validated, it is not clear whether it still outputs realistic reactions:
for example, a change in the solution rate could just be the result of new unrealistic reactions being output by the model.
Second, even disregarding model quality,
comparing search algorithms with different single-step models is essentially comparing
two algorithms in different environments, which may invalidate comparisons.
We think a better practice in this aspect is training a policy model to \emph{re-rank}
the top-$k$ outputs of a fixed single-step model without changing the set of feasible reactions.
This allows for meaningful improvement while still keeping the same accuracy guarantees and comparability of using the original single-step model.
We note that this strategy was recently used by \citet{liu2023retrosynthetic}.

\refstepcounter{multi-step-recs}\label{msr:success-rate}
\subsubsection*{Pitfall M2: using search success rate alone to compare single-step models}

\looseness=-1
Some works~\citep{segler2018planning,hassen2022mind, torren2023models} run search using various single-step models and use the success of such search to rank the models themselves. While we agree that single-step models should be benchmarked as part of search, inferring that a model is better solely because it allows for finding more routes can lead to flawed conclusions: an overly permissive single-step model may yield many routes simply because it lets search make unrealistic retrosynthetic steps, as demonstrated in contrasting experiments~\citep{segler2018planning}. Instead, success rate should be treated as an initial metric; a final determination of whether one end-to-end retrosynthesis pipeline is better than another is only possible if the quality of routes found is properly assessed. Outside of quality assessment or actually running synthesis, this could also be achieved using a forward prediction or feasibility model; however, training such models in a generalizable way is so far an underexplored research direction (see also M\ref{msr:chemist-eval}).

\refstepcounter{multi-step-recs}\label{msr:limit}
\subsubsection*{Best practice M3: carefully choose how search is capped if varying the single-step model}
Existing works differ in how search experiments are limited: some use number of calls to the reaction model~\citep{tripp2022reevaluating}, while others combine this with a wall-clock time limit~\citep{hassen2022mind}. Capping the number of model calls is a reliable choice if the single-step model is kept fixed; however, varying the single-step model can lead to some models being allocated vastly more \textit{resources} (e.g. time) than others~\citep{torren2023models}. This may be justified if one believes the model speed is subject to change, and that perhaps all compared models can be optimized to eventually take a similar amount of time per call, but in the absence of such belief we recommend limiting search using a measure that treats the algorithm as a black-box (e.g. wall-clock time or memory consumption), as such approach also more directly reflects downstream use in CASP systems.

\refstepcounter{multi-step-recs}\label{msr:cache}
\subsubsection*{Best practice M4: cache reaction model calls}

\looseness=-1
If the same molecule is encountered twice during search, a naive implementation will call the reaction model twice.
As calling the reaction model is expensive, a well-engineered CASP
system would clearly \emph{cache} the outputs of the reaction model to avoid duplicate computation.
Therefore, we believe it is best practice to use a cache for the single-step model when evaluating multi-step algorithms.
This may sound like a minor implementation detail, but it actually has a significant impact on the evaluation: often large sub-trees can occur in multiple places during search;\footnote{
For example, if the reactions $M\rightarrow A + B$ and $M\rightarrow A + C$ are possible,
then any subsequent reactions on molecule $A$ will be repeated multiple times.}
without a cache, expanding each occurrence of these subtrees will count against an algorithm's time budget,
whereas with a cache these expansions are effectively free~\citep{tripp2022reevaluating}.

\refstepcounter{multi-step-recs}\label{msr:diversity}
\subsubsection*{Best practice M5: evaluate route diversity}
\looseness=-1
While previous works emphasize finding a single synthesis route quickly,
because outputs of CASP programs may not work in the wet lab it is preferable to return
multiple routes, and that these routes be \emph{diverse}.
Put another way, once an algorithm is able to find a single route,
it is desirable to evaluate its ability to find additional ones
which differ from the one already found.
There are many ways to measure diversity, but we think that a good diversity metric
must be monotonic with respect to input routes (otherwise algorithms could be penalized for finding more routes).
One such metric is the packing number, also called \#Circles~\citep{xie2023much},
which can be instantiated as the number of synthesis routes with no overlapping reactions.\footnote{Finding the largest set of non-overlapping routes is equivalent to the set packing problem, which is NP-hard~\citep{karp2010reducibility}. In practice there is no need to compute it exactly, and a heuristic approximation is sufficient.}

\refstepcounter{multi-step-recs}\label{msr:chemist-eval}
\subsubsection*{Best practice M6: perform quality assessment by chemists}
Finally, we recommend practitioners perform a controlled qualitative assessment of the discovered routes by expert chemists, similarly to as in prior work~\citep{segler2018planning}. For example, this could be an A/B-test, and should be supported by adequate statistical analysis. Qualitative testing has the potential to catch many pitfalls, including (but not limited to) most of those described above. Additionally, it can capture poor synthesis strategies e.g. repetitions of similar steps, redundant (de)protection chemistry, or poor choice of linear vs convergent synthesis routes, which are difficult to spot with computational metrics. Expert assessment is laborious, however we argue it provides a middle ground compared to much more expensive wet-lab experiments. While the ultimate goal of synthesis planning tools is use in the wet-lab, we argue that wet-lab validation should not become a mandatory metric, since it would exclude computational researchers, and is very challenging to perform in an end-to-end controlled setting.  As a proxy, \citet{genheden2022paroutes} proposed to use recall of and similarity to known routes. 

\section{Syntheseus}
\label{sec:syntheseus}

To encourage and promote the principles and practices discussed in Section~\ref{sec:discussion}, we built a benchmarking library called \textsc{syntheseus}.
\textsc{Syntheseus} is designed to be a platform for researchers developing methods for retrosynthesis,
rather than a specific set of models or tasks.
Currently, there is no generic package for retrosynthesis evaluation,
forcing researchers to either write evaluation code themselves (which can be subtly inconsistent with prior work)
or directly copy code from prior works (which have not followed the best practices from Section~\ref{sec:discussion}).
\textsc{Syntheseus} provides an end-to-end retrosynthesis pipeline which is modular and extensible for both
novel single-step models and novel search algorithms;
this allows researchers to plug their methods into a well-tested framework which implements best practice by default.
We highlight key features of \textsc{syntheseus} below,
but refer the reader to \href{https://github.com/microsoft/syntheseus}{github.com/microsoft/syntheseus}
for an in-depth look into its API.

\subsection{Unrestricted single-step model development}
\textsc{Syntheseus} uses a minimal standard interface to interact with single-step models.
This enables users to build their models separately from \textsc{syntheseus}
and integrate them by writing a thin wrapper,
allowing \textsc{syntheseus} to evaluate and use all single-step models in a consistent way.
Furthermore,
as the framework controls the inputs and outputs of the wrapped model, it automatically prevents ``cheating'' in the form of relying on atom mappings (S\ref{ssr:data-leakage}),
takes care of post-processing the outputs when evaluating accuracy (S\ref{ssr:post-processing}), provides the ability to measure precision via round tripping with a forward model (S\ref{ssr:precision-recall}),
and measures inference time (S\ref{ssr:inference-time}).
When used in multi-step search, \textsc{syntheseus} also automatically performs caching (M\ref{msr:cache}).

\subsection{Separation of components in multi-step search}
\textsc{Syntheseus} cleanly separates the various components of a multi-step search algorithm:
the single-step model, set of purchasable molecules, the search graph, and search heuristics (policies and value functions).
This makes it easy to change one part of a CASP program and see the effect:
for example, run MCTS with two different single-step models,
or run Retro* with two different sets of purchasable molecules.

\subsection{Detailed metrics for multi-step search}
In addition to tracking whether a synthesis route has been found,
\textsc{syntheseus} also tracks \emph{when} it has been found
using several different time measures (wallclock time, number of calls to the reaction model), making it easy to track the performance of an algorithm over time.
\textsc{Syntheseus} also implements several diversity metrics (M\ref{msr:diversity}), and provides visualization tools to allow the routes to be inspected by researchers or expert chemists (M\ref{msr:chemist-eval}).

\section{Experiments: re-evaluation of existing methods}
\label{sec:experiments}

We use \textsc{syntheseus} to re-evaluate many existing single-step models in conjunction with popular search algorithms, providing a holistic view of the existing methods, and in many instances \emph{correcting the numbers from the literature}. We did not re-implement any of the models, and used open-source codebases wrapped into \textsc{syntheseus}'s single-step model interface, demonstrating its flexibility. Crucially, \emph{results in this paper were produced by our evaluation framework with no numbers copied from previous work}, which ensures a fair comparison immune to many issues discussed in Section~\ref{sec:discussion}. We note that the main purpose of this study is to showcase the framework, and not to extensively tune the models, which is left to future work. 

\subsection{Single-Step}

\subsubsection{Datasets}

As a starting point we use the USPTO-50K dataset~\citep{schneider2016s} split by~\citet{dai2019retrosynthesis}, as all of the models we consider report results on this dataset, allowing us to contrast the published numbers with ones obtained from our re-evaluation. There is a newer version of this dataset available~\citep{lin2022improving}, but since our aim is to correct the existing results, we focus on the more established version and leave using the newer one for future work. 
Moreover, USPTO-50K is a small dataset, and it may not be representative of the full data distribution. Thus, we also use the proprietary Pistachio dataset~\citep{mayfield2017pistachio} (more than 15.6M raw reactions; 3.4M samples after preprocessing), and evaluate out-of-distribution generalization of the model checkpoints trained on USPTO-50K. To the best of our knowledge this has not been done before; while some works also make use of Pistachio~\citep{jiang2023learning}, it is rather used as a \emph{pretraining} dataset before fine-tuning on USPTO. 
As most researchers do not have access to Pistachio, by reporting generalization we aim to gain insight into how USPTO-trained models work across a wider input distribution they may be exposed to during multi-step search. We performed extensive cleaning and preprocessing on the Pistachio data to ensure the test set is of sufficient quality, and also to limit overlap between the USPTO training set and Pistachio test set; see Appendix~\ref{appendix:pistachio} for the details of our preprocessing pipeline.

\begin{figure*}[t]
\vskip0.3cm
\centering
\includegraphics[height=4.95cm]{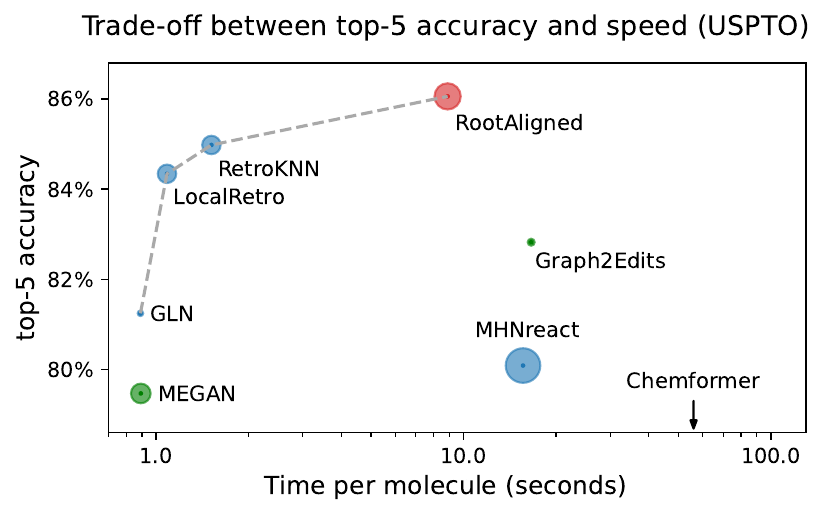}
\includegraphics[height=4.95cm]{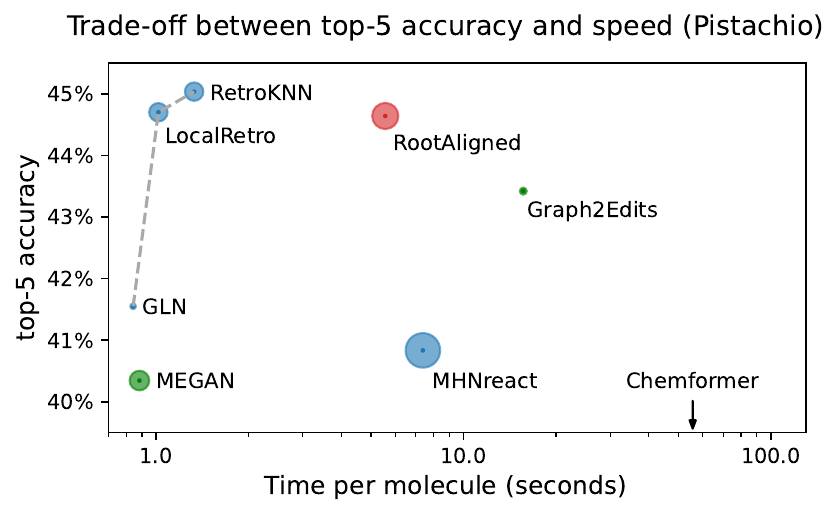}
\caption{
Trade-off between top-5 accuracy and inference speed. Circle area is proportional to the number of parameters; color denotes whether a model uses reaction templates (blue), generates a sequence of graph edits (green)  or produces the output SMILES from scratch (red). Dashed gray line shows the Pareto front (best result for any time budget). Exact results for Chemformer are not shown as they fall below the plot boundary. We show in-distribution results on USPTO-50K (left) and out-of-distribution generalization on Pistachio (right).}
\label{fig:single-step}
\end{figure*}

\subsubsection{Models}

We re-evaluate established single-step models where either the code is publicly available (Chemformer~\citep{irwin2022chemformer}, GLN~\citep{dai2019retrosynthesis}, Graph2Edits~\citep{zhong2023retrosynthesis}, LocalRetro~\citep{chen2021deep}, MEGAN~\citep{sacha2021molecule}, MHNreact~\citep{seidl2021modern} and RootAligned~\citep{zhong2022root}) or we were able to obtain it from the authors (RetroKNN~\citep{xie2023retroknn}). We omit Dual-\{TB,TF\}~\citep{sun2020energy} and E-SMILES~\citep{xiong2023improve} as we have no access to the code; even though these models reported promising performance, we were unable to verify it under our framework. For all models we used the provided checkpoint if one using the right data split was available, and trained a new model using the original training code otherwise.
We used the original implementations adapted to our shared interface.\footnote{Adapting MHNreact we found that its use of multiprocessing was suboptimal; our wrapped version partially fixes this and is more performant than the original code.}

\subsubsection{Metrics}
\looseness=-1
We compute top-$k$ accuracy for $k$ up to $50$ and Mean Reciprocal Rank (MRR). It is not clear what value of $k$ is the most relevant metric to consider, but given the target use of single-step models in search, it is desirable for $k$ to be roughly similar to the expected or desired breadth of the search tree (number of children visited for a typical internal node); thus, $k = 1$ would be too narrow. Typically, values beyond $k = 50$ are not reported, as models tend to saturate past this point. Several CASP programs also restrict the expansion beyond the top-50~\citep{segler2018planning, genheden2020aizynthfinder}. We highlight $k = 5$ as a reasonable middle-ground and defer extended results to Appendix~\ref{appendix:single-step-results}.

\subsubsection{Setup}

We queried all models for $n = 100$ outputs (see Appendix~\ref{appendix:multiple-single-step-results} for a discussion on how obtaining multiple results is handled for different model types).
Note that we measure top-$k$ only up to $k = 50$ but set $n > k$ to account for deduplication.
We used a fixed batch size of $1$ for all models.
While all models could easily handle larger batches, batch size used during search typically cannot be set arbitrarily, and in most cases it is equal to $1$ as usually search is not parallelized.
Thus, speed under batch size of $1$ directly translates to the maximum number of model calls that can be performed during search with a fixed time budget.
All inference time measurements used the same Microsoft Azure compute nodes with a single V100 GPU.

\subsubsection{Results}

\def\multistepwidthleft{7.0cm}
\def\multistepwidthright{6.776cm}

\begin{figure*}[h!]
\vskip0.3cm
\centering
\includegraphics[width=\multistepwidthleft]{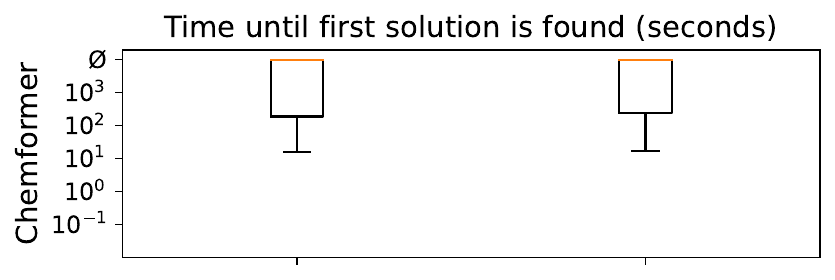}
\includegraphics[width=\multistepwidthright]{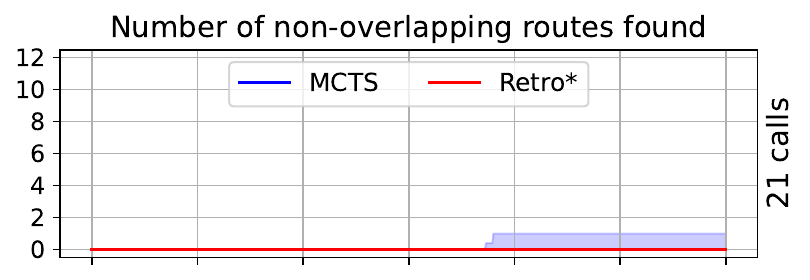}\\
\includegraphics[width=\multistepwidthleft]{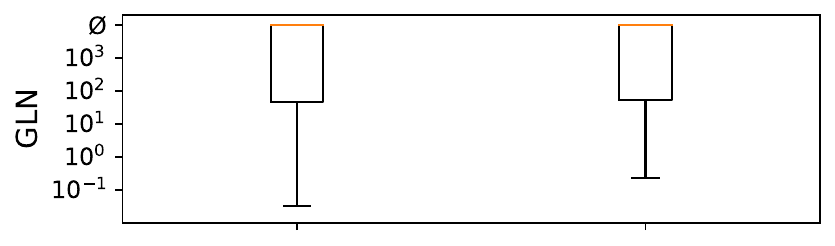}
\includegraphics[width=\multistepwidthright]{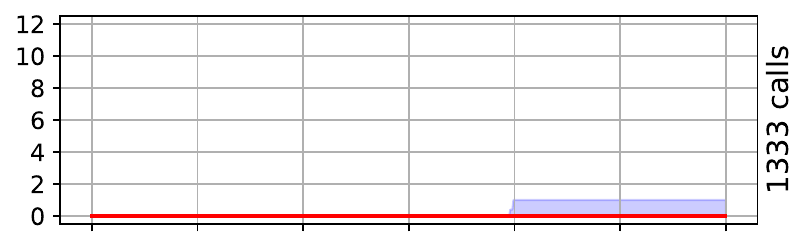}\\
\includegraphics[width=\multistepwidthleft]{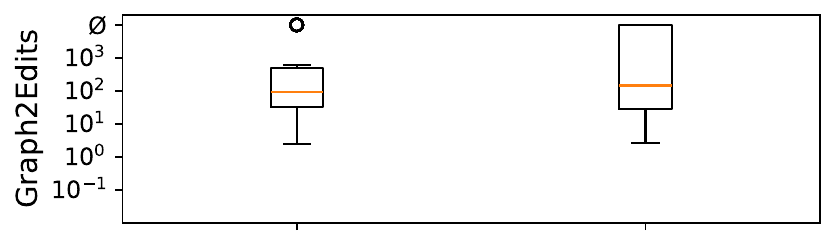}
\includegraphics[width=\multistepwidthright]{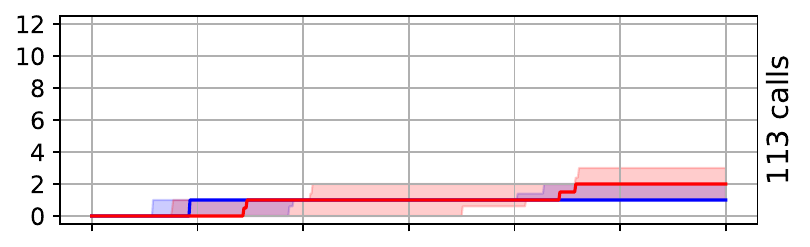}\\
\includegraphics[width=\multistepwidthleft]{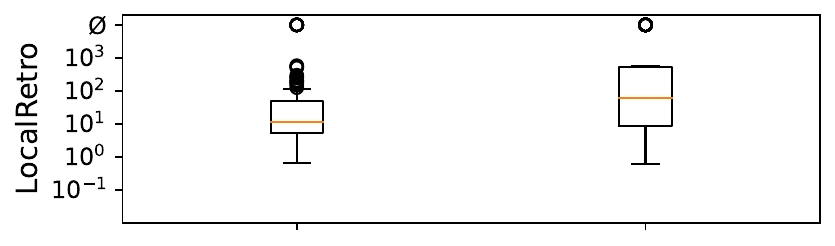}
\includegraphics[width=\multistepwidthright]{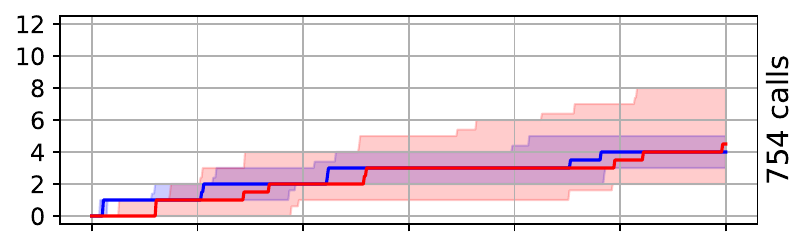}\\
\includegraphics[width=\multistepwidthleft]{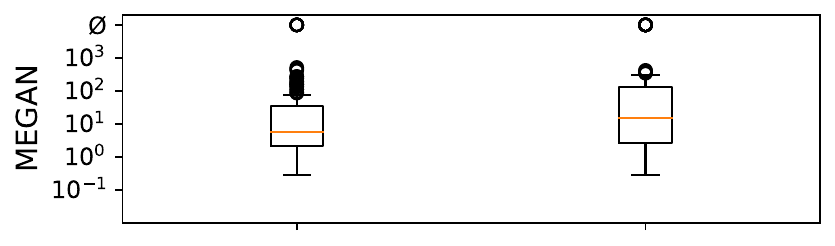}
\includegraphics[width=\multistepwidthright]{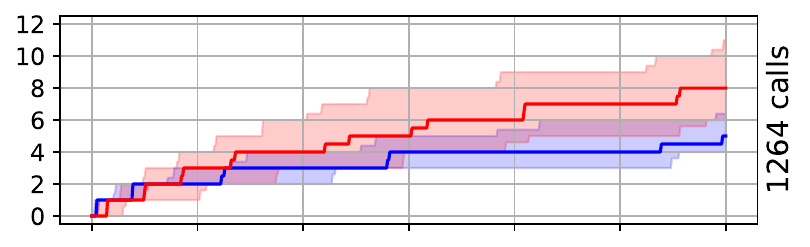}\\
\includegraphics[width=\multistepwidthleft]{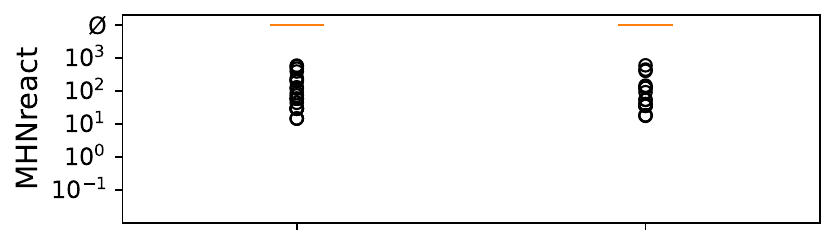}
\includegraphics[width=\multistepwidthright]{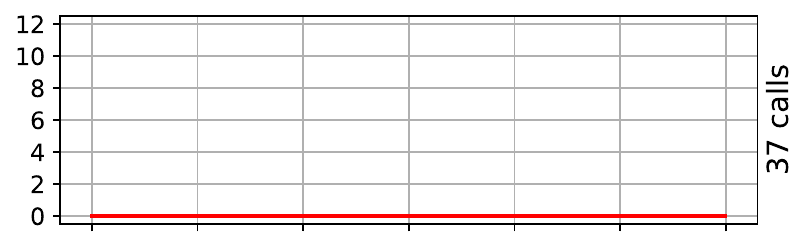}\\
\includegraphics[width=\multistepwidthleft]{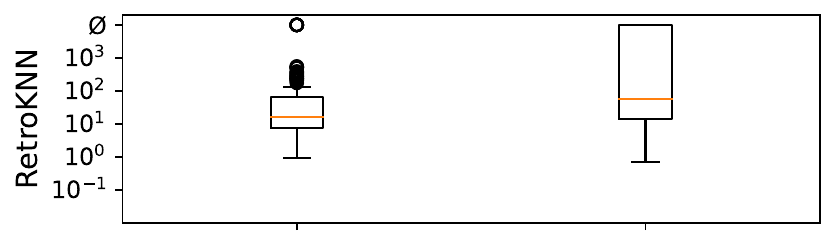}
\includegraphics[width=\multistepwidthright]{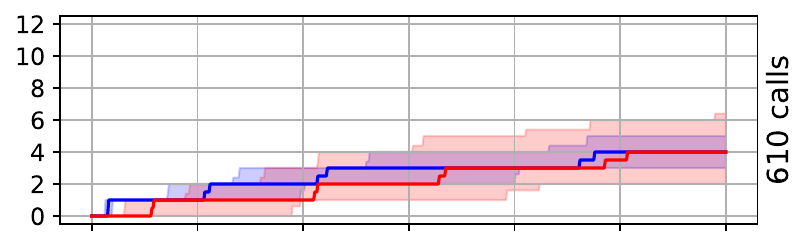}
\includegraphics[width=\multistepwidthleft]{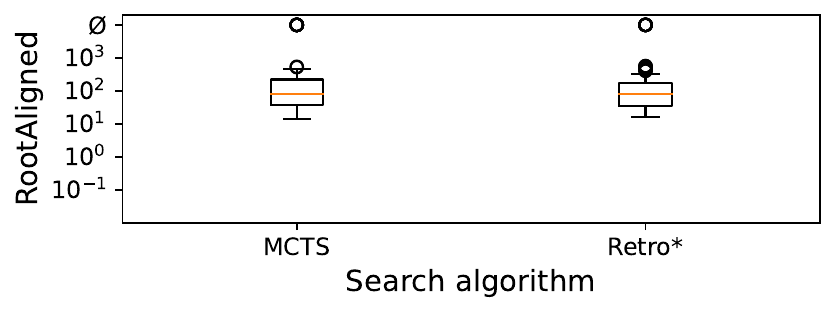}
\includegraphics[width=\multistepwidthright]{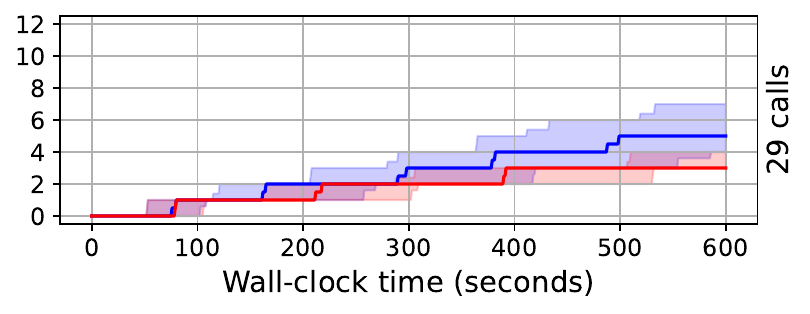}
\caption{Multi-step search results on the Retro* Hard target set with different single-step models. \textbf{Left:} Time until first solution was found (or $\emptyset$ if a molecule was not solved).
Orange line represents the median, box represents 25th and 75th percentile, whiskers represent 5th and 95th percentile, points outside this range are shown as dots.
\textbf{Right:} Approximate number of non-overlapping routes present in the search graph (tracked over time and aggregated across target molecules). Solid line represents the median, shaded area shows the 40th and 60th percentile. On the right hand side we note the average number of calls made by the model within the allotted time limit.}
\label{fig:multi-step}
\end{figure*}

\looseness=-1
We present top-5 accuracy results on both datasets in Figure~\ref{fig:single-step}.
First, we note that two of the models (RootAligned, Chemformer) predict the reactants SMILES from scratch using a Transformer decoder~\citep{vaswani2017attention}, while the other models predict the graph rewrite to apply to the product. Across datasets and metrics, models of the former type tend to be slower, and while they show good performance in top-1 accuracy, they are surpassed by the graph-transformation-based models for higher $k$. We hypothesize that, due to more explicit grounding in the set of transformations occurring in training data, transformation-based models tend to produce a more complete coverage of the data distribution.
Second, many of the USPTO-50K results we report are better than the numbers from the literature (see Appendix~\ref{appendix:single-step-results} for a detailed breakdown), especially in terms of top-$k$ accuracy for $k > 1$, which is affected by deduplication. This also changes some of the model rankings, e.g. LocalRetro was originally reported to have a better top-1 accuracy than GLN, but we find that to not be the case.
Surprisingly, model ranking on USPTO-50K transfers to Pistachio quite well, although all results are substantially degraded, e.g. in terms of top-50 accuracy all models still fall below $55\%$, compared to nearly $100\%$ on USPTO.
While for template-based models this is a result of insufficient coverage, we note that some of the models tested here are template-free, and yet they fail to generalize better than their template-based counterparts (this is similar to the findings of~\citet{tu2022retrosynthesis}).
To further ground our Pistachio results, we note that~\citet{jiang2023learning} report 66.1\% top-5 accuracy when training on Pistachio directly (compared to our transfer results of up to 45\%); however, these values are not fully comparable due to differences in preprocessing.
Finally, RetroKNN is best or close to best on all metrics on both datasets, while also being one of the faster models in our re-evaluation.

\subsection{Multi-Step}

\looseness=-1
We also ran search experiments combining various single-step models and search algorithms.
As our primary objectives are to outline good practices, correct established numbers, and showcase \textsc{syntheseus}, we only show preliminary multi-step results; since all search algorithms also have hyperparameters which require extensive tuning, we leave a further determination of which end-to-end pipeline is best to future work building on top of our framework. 
It is worth noting that the single-step models considered here use various (usually conflicting) versions of deep learning frameworks and other libraries, yet due to minimalistic dependencies \textsc{syntheseus} can be combined with any of them.

\subsubsection{Setup}

We followed one of the experimental setups from~\citet{tripp2022reevaluating} and used the 190 target molecules from Retro* Hard~\citep{chen20retrostar}, together with the $23\,081\,629$ building block molecules from the eMolecules set.
We combined each of the eight single-step models with two search algorithms: an MCTS variant and Retro*~\citep{chen20retrostar}. All runs were limited to $10$ minutes per molecule.
Search graphs for MCTS and Retro* were capped to a maximum depth of $20$ and $10$, respectively, which for MCTS translates to routes containing at most $20$ reactions overall, while for Retro* routes with a depth of at most $5$ reactions (each reaction uses two nodes in an AND/OR graph) but potentially many more reactions in total for convergent routes.
Our single-step model wrappers expose the underlying output probabilities, which are used by both algorithms to guide the search.
To ensure a fair comparison, the hyperparameters of each algorithm-model combination were tuned separately (see Appendix~\ref{appendix:search-tuning} for the exact procedure; qualitatively this step was especially important for MCTS).

\subsubsection{Results}

We show the results in Figure~\ref{fig:multi-step}, tracking when the first solution was found as well as the maximum number of non-overlapping routes that can be extracted from the search graph.
For most models (all apart from Chemformer, GLN and MHNreact), both search algorithms are able to find several disjoint routes for the majority of targets.
Notably, RootAligned obtains promising results despite making less than 30 calls on average (due to its high computational cost). Interestingly, we also find that the difference between Retro* and our MCTS reimplementation is small, which is in line with the results by \citet{genheden2022paroutes}, but contrasts the results from the original Retro* paper~\citep{chen20retrostar}. 
However, as discussed in Section~\ref{sec:discussion-multi-step}, these results should not be treated as a final comparison of the models and rather serve as a starting point for future research.

\section{Conclusion and Future Work}
\label{sec:summary}

In this paper we presented an analysis of pitfalls and best practices for evaluating retrosynthesis programs (Section~\ref{sec:discussion}),
a synthesis planning software package called \textsc{syntheseus} to help researchers benchmark their methods following these best practices (Section~\ref{sec:syntheseus}),
and used \textsc{syntheseus} to re-evaluate many existing models and algorithms (Section~\ref{sec:experiments}).
These results ``set the record straight'' regarding the performance of existing algorithms,
and the standardized evaluation protocol of \textsc{syntheseus} can ensure that future works do not continue to make the same mistakes.
We encourage members of the community to contribute new models, algorithms, and metrics to \textsc{syntheseus}
(see maintenance plan in Appendix~\ref{appendix:maintenance}).

Despite this, several important issues remain in the field, which we plan to resolve with \textsc{syntheseus} in future iterations.
As we argue in Section~\ref{sec:discussion}, existing metrics of recall (S\ref{ssr:precision-recall}) and solve rate (M\ref{msr:single-step-model}/M\ref{msr:success-rate}) are not ideal for comparing arbitrary end-to-end retrosynthesis pipelines.
Assuming evaluation by chemists (M\ref{msr:chemist-eval}) is not possible,
we believe the most plausible substitute is to develop improved 
forward reaction prediction or reaction feasibility models to estimate whether reactions will succeed.
If such models were used post-hoc (not available during training or search),
they could be used to evaluate the precision of single-step models (resolving S\ref{ssr:precision-recall})
and assign a feasibility score to entire routes (resolving M\ref{msr:single-step-model}/M\ref{msr:success-rate}).
We designed \textsc{syntheseus} with this in mind and have a clear way to support feasibility models in both single-step and multi-step evaluation.
However, how to train a high-quality feasibility model is an open research question which we leave to future work.
Finally, the lack of information on reaction conditions, required quality of the starting materials, required equipment,
and purification,
is a significant barrier to actually executing the synthesis plans from CASP systems which \textsc{syntheseus} does not address. We encourage the community to work together with us on these challenges in the future.

\section*{Author Contributions}

The single-step reaction prediction workstream was led by Krzysztof, who built the majority of the codebase, ran the experiments, and integrated several models (LocalRetro, GLN and Graph2Edits). Others contributed to specific components: Guoqing (parts of evaluation pipeline, Pistachio data preprocessing, integrating RootAligned), Megan (parts of evaluation pipeline, integrating MEGAN and Chemformer), Shufang (integrating RetroKNN), Piotr (analysing the utility of Chemformer as a feasibility model), Philipp (integrating MHNreact) and Marwin (Pistachio data analysis and preprocessing, template extraction). The multi-step search workstream was led by Austin, who implemented all of the search algorithms and ran initial experiments, advised jointly by Krzysztof and Marwin. Experiments with search algorithms were continued by Krzysztof, who performed additional tuning and produced the final results. Writing was done together by Krzysztof and Austin, with help from Marwin and Guoqing, as well as comments from the other authors. Both workstreams were mentored by Marwin.

\section*{Acknowledgments}

We would like to thank Hannes Schulz, Maik Riechert, Ran Bi, as well as the larger Microsoft Research team, for help with computing infrastructure that made this work possible. We would further like to acknowledge Elise van der Pol and Jose Garrido Torres for helpful discussions.

\bibliography{main}
\bibliographystyle{icml2024}

\clearpage
\appendix
\onecolumn

\section{Pistachio Preprocessing} \label{appendix:pistachio}

The raw Pistachio data (version 2023Q2, released in June 2023) contained $15\,684\,711$ raw reactions; however, this included many duplicates, outliers (e.g. reactions with extremely large products), and potentially samples that are erroneous.
To ensure the test data is of high quality, we performed careful filtering and processing of the raw Pistachio data. We applied the following steps in order:

\begin{itemize}
    \item Remove duplicate reactions.
    \item Remove reactions with more than $4$ reactants.
    \item Compute occurrence count of each product molecule across the dataset (counting individual products in multi-product reactions separately). For every reaction with products $[p_1, ..., p_m]$ (including reactions with a single product i.e. $m = 1$), remove all side products. Product $p_i$ is considered a side product if it either has less than $5$ atoms, or appears at least $1000$ times across the dataset. The latter condition allows us to remove common side products, which may have $5$ or more atoms but are still uninteresting. Retain only those reactions where exactly one $p_i$ remained after this procedure (i.e. those with a well-defined main product).
    \item Remove reactions where the (now unique) product has more than $100$ atoms.
    \item Remove reactions where the ratio of the number of reactant atoms to the number of product atoms exceeds $20$.
    \item Remove reactions where the product appears as one of the reactants.
    \item  Refine reactions by removing the atom mapping numbers that appear only on one side.
    \item Remove reactions that have double-mapped atoms on either the main product side or the reactants side, or those that lack atom mapping numbers entirely.
    \item Refine reactions by removing reactants that do not contribute atoms to the product.
\end{itemize}

We chose the processing steps above such that we exclude erroneous reactions, extreme outliers (i.e. those that are either very large or have an extreme imbalance between the size of the reactants and the size of the product), and reactions with no clearly defined main product. These processing steps (and the particular constants used therein) were informed by expert qualitative analysis of the reactions, as well as practical considerations.
For example, we found that several very large outliers in raw Pistachio data seem to cause \texttt{rdchiral}'s~\citep{coley2019rdchiral} template extraction routines to hang; however, these reactions did not survive our filtering.
Further discussion on preprocessing chemical reaction data for use in Deep Learning can be found in~\citet{wigh2023orderly}.

After the preprocessing we obtained $3\,445\,833$ single-product samples, which we grouped by their product, and split into train, validation and test sets following a 90/5/5 ratio, making sure the groups of samples with the same product are placed into the same fold. We used a random split, except for those products which were found in USPTO-50K data; in those cases, we attempt to place the corresponding group of samples in the same fold as it appears in USPTO (this limits overlap between training set of one dataset and test set of another, which could distort our generalization results). As the USPTO-50K split from~\citet{dai2019retrosynthesis} contains a small amount of product overlap between folds, this process of ensuring a ``compatible'' Pistachio split was imperfect; the product overlap between USPTO-50K training set and Pistachio test set is non-zero, but it is negligibly small.
Note that an alternative approach to preventing overlap would be to completely remove USPTO products from Pistachio before splitting the dataset, but we did not want to artificially exclude (valuable) products present in USPTO-50K.
Finally, at the end we also removed a small number of reactions in all folds for which template extraction using \texttt{rdchiral} fails, as this often signifies that the atom mapping is erroneous.

In this work we use Pistachio solely for testing generalization, thus we only used the test fold, which we randomly subsampled to $20\,000$ samples for faster evaluation (note that this is still 4 times larger than the test set of USPTO-50K).
We described the full procedure to generate all folds to facilitate future work.

\clearpage

\section{Extended Single-step Results}\label{appendix:single-step-results}

\subsection{Trade-off Between Quantitative Performance and Speed}

\begin{figure*}[h]
\centering
\includegraphics[height=4.95cm]{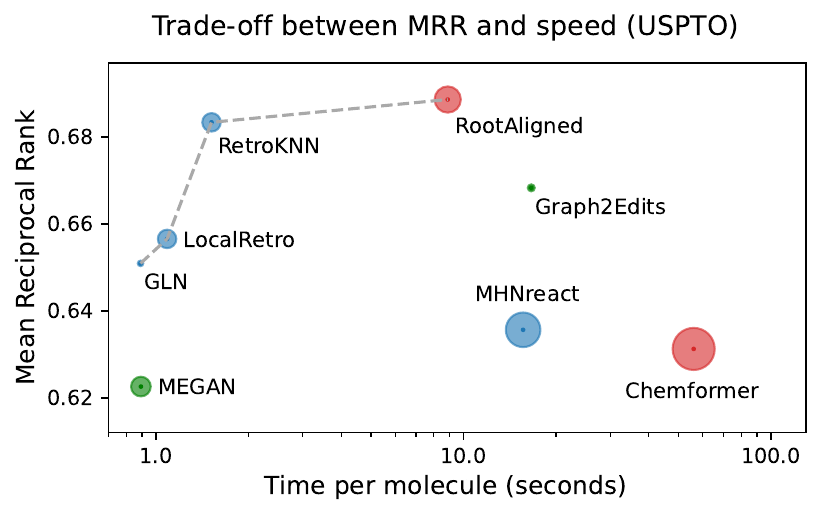}
\includegraphics[height=4.95cm]{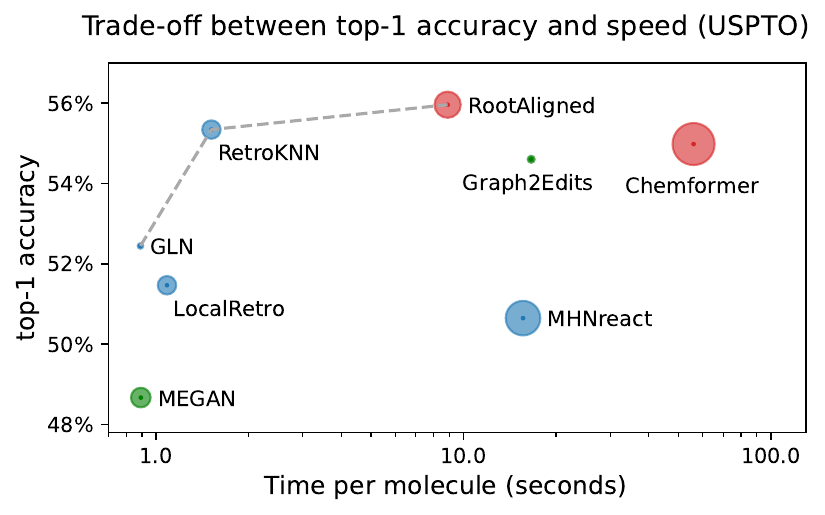}\\
\includegraphics[height=4.95cm]{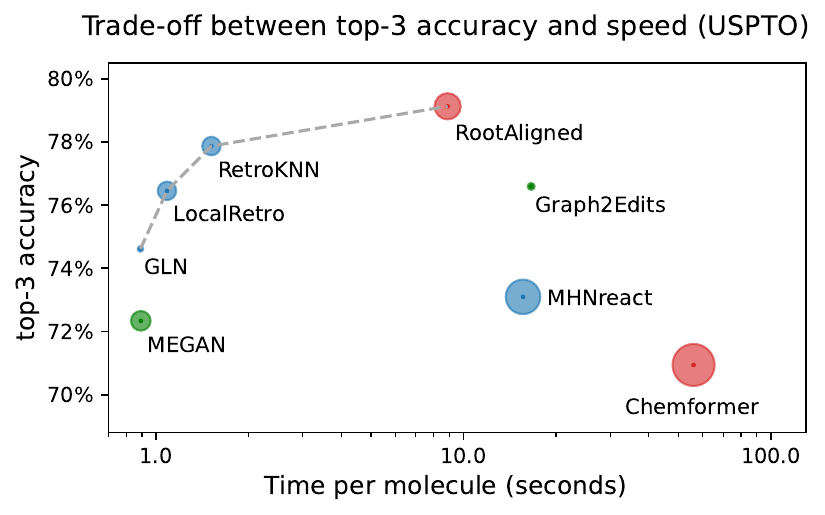}
\includegraphics[height=4.95cm]{figures/single_step/top_5_uspto.pdf}\\
\includegraphics[height=4.95cm]{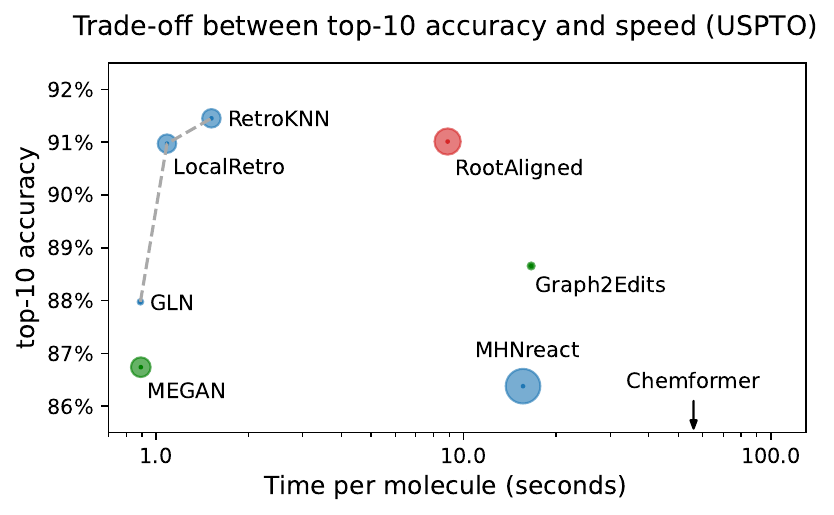}
\includegraphics[height=4.95cm]{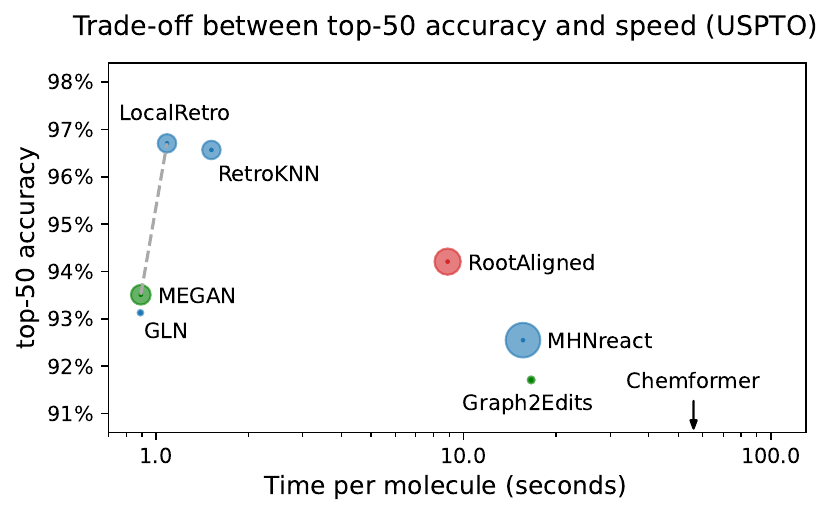}
\caption{Results on USPTO-50K in same format as Figure~\ref{fig:single-step}  but extended with top-1, top-3, top-10, top-50, and MRR. Plot for top-5 is reprinted here for convenience.}
\label{fig:single-step-uspto-extended}
\end{figure*}

\begin{figure}[h]
\centering
\includegraphics[height=4.95cm]{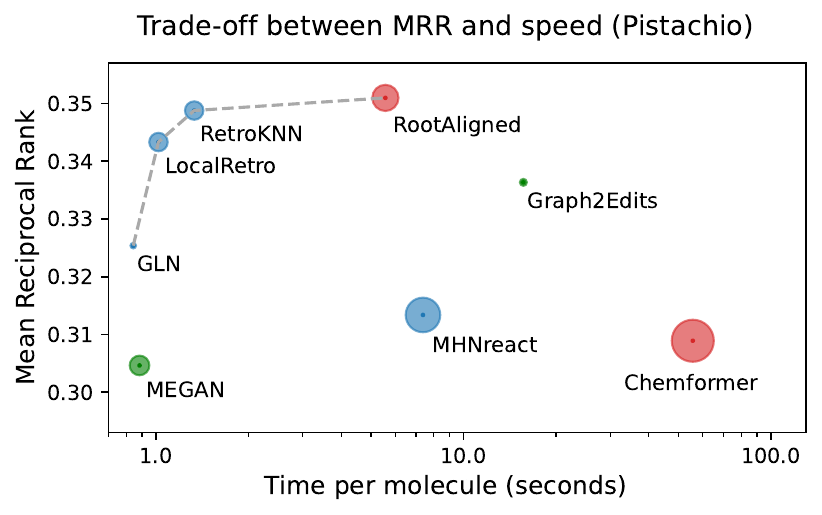}
\includegraphics[height=4.95cm]{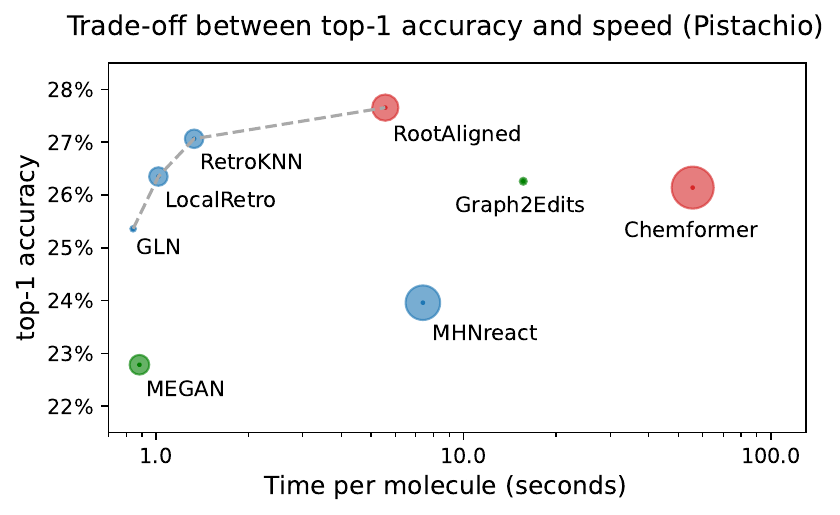}\\
\includegraphics[height=4.95cm]{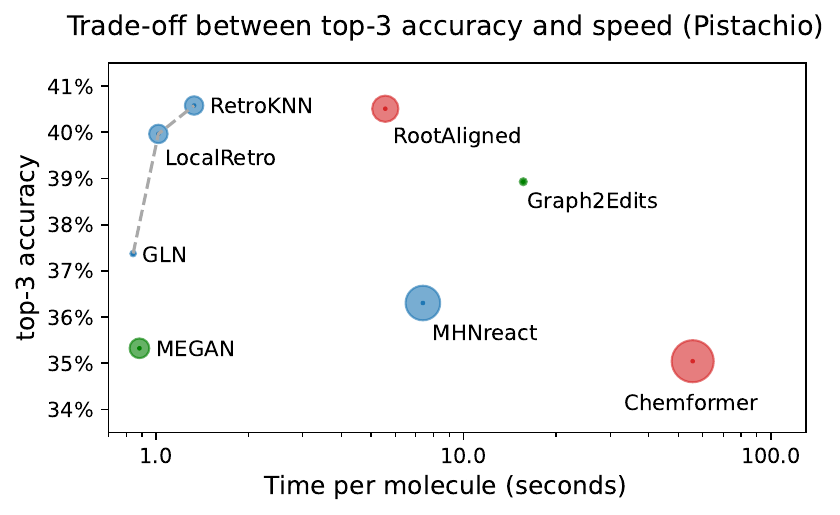}
\includegraphics[height=4.95cm]{figures/single_step/top_5_pistachio.pdf}\\
\includegraphics[height=4.95cm]{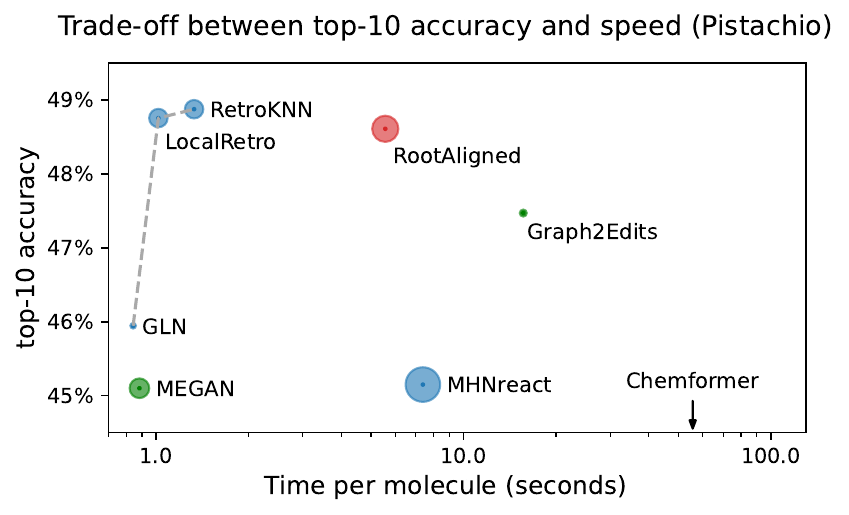}
\includegraphics[height=4.95cm]{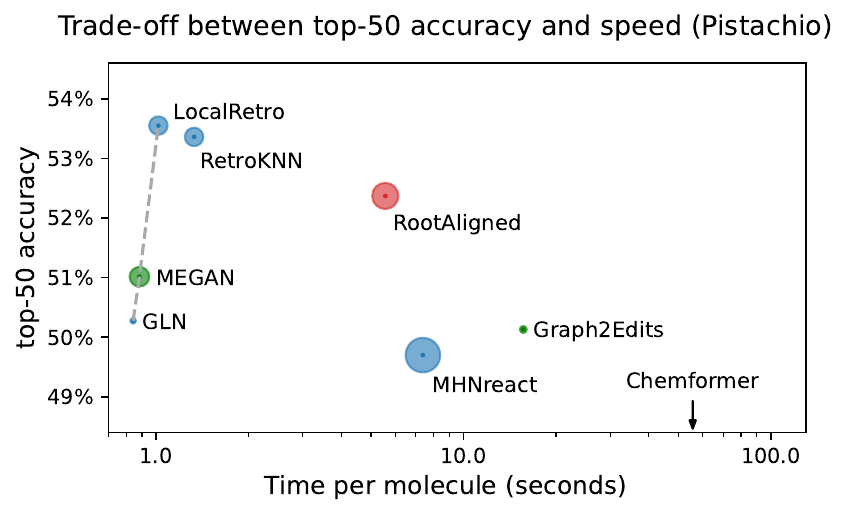}
\caption{Results on Pistachio in same format as Figure~\ref{fig:single-step} but extended with top-1, top-3, top-10, top-50, and MRR. Plot for top-5 is reprinted here for convenience.}
\label{fig:single-step-pistachio-extended}
\end{figure}

\clearpage

\subsection{Comparison with Published Results}

In Table~\ref{tab:single-step-uspto}, we present the results from Figure~\ref{fig:single-step-uspto-extended} in numeric form, as well as contrast them with the published numbers. For results produced with \textsc{syntheseus} we additionally investigate the effect of deduplication.

\definecolor{cadmiumgreen}{rgb}{0.0, 0.42, 0.24}
\newcommand{\increased}[1]{\textcolor{cadmiumgreen}{\underline{#1}}}
\newcommand{\decreased}[1]{\textcolor{red}{\underline{#1}}}

\begin{table}[h]
\caption{Results on USPTO-50K compared to the numbers reported in the literature. \textsc{Synth} denotes whether we used \textsc{syntheseus} to produce the result (as opposed to copying the published number, or, in case of LocalRetro with exact match, generating the number ourselves using authors' code), \textsc{D} denotes whether deduplication was performed (which in \textsc{syntheseus} is enabled by default, but can be turned off). We underline values that differ significantly from the previous row (at least $0.7\%$ for top-$k$ or $0.003$ for MRR), and use colors to distinguish whether the value is better (\textcolor{cadmiumgreen}{green}) or worse (\textcolor{red}{red}) than the row directly above.}
\vskip0.15in
\label{tab:single-step-uspto}
\centering
\begin{tabular}{lccccccccc}
\toprule
Model & \textsc{Synth} & \textsc{D} & top-1 & top-3 & top-5 & top-10 & top-50 & MRR \\
\midrule
Chemformer & \xmark &  & $54.3\%$ & - & $62.3\%$ & $63.0\%$ & - & -\\
 & \cmark & \xmark & \increased{$55.0\%$} & $67.8\%$ & \increased{$70.5\%$} & \increased{$72.5\%$} & $74.8\%$ & $0.6182$\\
 & \cmark & \cmark & $55.0\%$ & \increased{$70.9\%$} & \increased{$73.7\%$} & \increased{$75.4\%$} & \increased{$76.0\%$} & \increased{$0.6312$}\\
\midrule
GLN & \xmark &  & $52.5\%$ & $69.0\%$ & $75.6\%$ & $83.7\%$ & $92.4\%$ & -\\
 & \cmark & \xmark & $52.4\%$ & $68.8\%$ & $75.4\%$ & $83.5\%$ & $92.5\%$ & $0.6262$\\
 & \cmark & \cmark & $52.4\%$ & \increased{$74.6\%$} & \increased{$81.2\%$} & \increased{$88.0\%$} & $93.1\%$ & \increased{$0.6509$}\\
\midrule
Graph2Edits & \xmark &  & $55.1\%$ & $77.3\%$ & $83.4\%$ & $89.4\%$ & $92.7\%$ & -\\
 & \cmark & \xmark & $54.6\%$ & \decreased{$76.4\%$} & \decreased{$82.6\%$} & \decreased{$88.5\%$} & \decreased{$91.7\%$} & $0.6672$\\
 & \cmark & \cmark & $54.6\%$ & $76.6\%$ & $82.8\%$ & $88.7\%$ & $91.7\%$ & $0.6683$\\
\midrule
LocalRetro & \xmark &  & $53.4\%$ & $77.5\%$ & $85.9\%$ & $92.4\%$ & $97.7\%$ & -\\
(exact match) & \xmark &  & \decreased{$52.0\%$} & \decreased{$75.5\%$} & \decreased{$83.4\%$} & \decreased{$90.0\%$} & \decreased{$95.7\%$} & -\\
 & \cmark & \xmark & $51.5\%$ & $75.6\%$ & $83.5\%$ & $90.6\%$ & \increased{$96.7\%$} & $0.6530$\\
 & \cmark & \cmark & $51.5\%$ & \increased{$76.5\%$} & \increased{$84.3\%$} & $91.0\%$ & $96.7\%$ & \increased{$0.6565$}\\
\midrule
MEGAN & \xmark &  & $48.1\%$ & $70.7\%$ & $78.4\%$ & $86.1\%$ & $93.2\%$ & -\\
 & \cmark & \xmark & $48.7\%$ & \increased{$71.9\%$} & $78.9\%$ & $86.0\%$ & $93.2\%$ & $0.6203$\\
 & \cmark & \cmark & $48.7\%$ & $72.3\%$ & $79.5\%$ & \increased{$86.7\%$} & $93.5\%$ & $0.6226$\\
\midrule
MHNreact & \xmark &  & $50.5\%$ & $73.9\%$ & $81.0\%$ & $87.9\%$ & $94.1\%$ & -\\
 & \cmark & \xmark & $50.6\%$ & \decreased{$73.1\%$} & \decreased{$80.1\%$} & \decreased{$86.4\%$} & \decreased{$92.6\%$} & $0.6356$\\
 & \cmark & \cmark & $50.6\%$ & $73.1\%$ & $80.1\%$ & $86.4\%$ & $92.6\%$ & $0.6356$\\
\midrule
RetroKNN & \xmark &  & $57.2\%$ & $78.9\%$ & $86.4\%$ & $92.7\%$ & $98.1\%$ & -\\
 & \cmark & \xmark & \decreased{$55.3\%$} & \decreased{$76.9\%$} & \decreased{$84.3\%$} & \decreased{$90.8\%$} & \decreased{$96.5\%$} & $0.6796$\\
 & \cmark & \cmark & $55.3\%$ & \increased{$77.9\%$} & \increased{$85.0\%$} & \increased{$91.5\%$} & $96.6\%$ & \increased{$0.6834$}\\
\midrule
RootAligned & \xmark &  & $56.3\%$ & $79.2\%$ & $86.2\%$ & $91.0\%$ & $94.6\%$ & -\\
 & \cmark & \xmark & $56.0\%$ & $79.1\%$ & $86.1\%$ & $91.0\%$ & $94.2\%$ & $0.6886$\\
 & \cmark & \cmark & $56.0\%$ & $79.1\%$ & $86.1\%$ & $91.0\%$ & $94.2\%$ & $0.6886$\\
\bottomrule
\end{tabular}
\end{table}

Focusing on the most significant differences between the results, we make the following observations:

\begin{itemize}
    \item Chemformer's results are improved when switching to \textsc{syntheseus}, and then further when turning on deduplication. The former could be explained by the fact that \textsc{syntheseus} removes invalid molecules, which Chemformer (as a SMILES-based model) can produce.
    \item GLN's published results match those obtained with \textsc{syntheseus} with no deduplication. However, its top-$k$ accuracies for $k > 1$ improve significantly with deduplication turned on.
    \item Graph2Edits' results are slightly worse than originally published, which may be explained by differences in hyperparameters such as the number of beams.
    \item LocalRetro (and by extension RetroKNN) used a relaxed notion of success, and we see that the results deteriorate significantly when using \textsc{syntheseus}. For LocalRetro, we additionally measured accuracy using authors' original code but replacing the relaxed match with an exact one (see row labelled with ``(exact match)''), which caused a similar drop in performance, confirming that the way of measuring accuracy is indeed responsible for the difference. Both LocalRetro and RetroKNN improve due to deduplication, but the final results still fall short of the originally reported numbers.
    \item MEGAN's published results improve slightly after moving to \textsc{synthesues}, and then there is a small further improvement from deduplication. We hypothesize the former might be a result of retraining the model (while the authors did release a checkpoint trained on USPTO-50K, our analysis seemed to indicate that model used a different data split for training, as the performance on our USPTO-50K test set was unrealistically high).
    \item MHNreact's results are not affected by deduplication, but the numbers we obtain with \textsc{syntheseus} are worse than those originally published; this may be explained by either the fact that we retrained the model or implementation details.
    \item RootAligned's published results closely match those obtained with \textsc{syntheseus} and are unaffected by deduplication, showing this model likely already conforms to many of the best practices from Section~\ref{sec:discussion}.
\end{itemize}

Next, in Table~\ref{tab:single-step-pistachio} we present the exact numbers corresponding to the results from Figure~\ref{fig:single-step-pistachio-extended}. However, here we cannot compare to published results, as to the best of our knowledge these are not available.

\begin{table}[h]
\caption{Generalization results on Pistachio in numeric form.}
\vskip0.15in
\label{tab:single-step-pistachio}
\centering
\begin{tabular}{lcccccc}
\toprule
Model & top-1 & top-3 & top-5 & top-10 & top-50 & MRR \\
\midrule
Chemformer & $26.1\%$ & $35.0\%$ & $37.1\%$ & $38.4\%$ & $39.1\%$ & $0.3089$\\
GLN & $25.4\%$ & $37.4\%$ & $41.5\%$ & $45.9\%$ & $50.3\%$ & $0.3254$\\
Graph2Edits & $26.3\%$ & $38.9\%$ & $43.4\%$ & $47.5\%$ & $50.1\%$ & $0.3363$\\
LocalRetro & $26.4\%$ & $40.0\%$ & $44.7\%$ & $48.8\%$ & $53.5\%$ & $0.3433$\\
MEGAN & $22.8\%$ & $35.3\%$ & $40.3\%$ & $45.1\%$ & $51.0\%$ & $0.3046$\\
MHNreact & $24.0\%$ & $36.3\%$ & $40.8\%$ & $45.1\%$ & $49.7\%$ & $0.3134$\\
RetroKNN & $27.1\%$ & $40.6\%$ & $45.0\%$ & $48.9\%$ & $53.4\%$ & $0.3488$\\
RootAligned & $27.7\%$ & $40.5\%$ & $44.6\%$ & $48.6\%$ & $52.4\%$ & $0.3510$\\
\bottomrule
\end{tabular}
\end{table}

\section{Obtaining Multiple Results from Single-step Models}\label{appendix:multiple-single-step-results}

During evaluation, we need to obtain $n$ results for a given input. It is worth noting that the value of $n$ is used differently depending on model type: models based on templates and local templates (GLN, LocalRetro, MHNreact and RetroKNN) first process the input and then apply the templates until $n$ results are obtained, while models that employ a sequential auto-regressive decoder (Chemformer, Graph2Edits and MEGAN) use beam search with $n$ beams. These two approaches lead to different scaling, as in the former case the bulk of the computation is amortized and does not scale with $n$, while in the latter case the entire procedure scales with $n$ essentially linearly. Finally, the RootAligned model is a special case, as it uses a combination of beam search and test-time data augmentation; scaling up either of these hyperparameters increases inference time and number of results, but the right balance between them requires careful tuning. In our work we used the default settings ($20$ augmentations, $10$ beams) which correspond to a maximum of $20 \cdot 10 > n$ results being generated (recall that $n = 100$).

\section{Search Algorithms Hyperparameter Tuning}
\label{appendix:search-tuning}

To ensure a fair comparison, we tuned the hyperparameters of both MCTS and Retro* separately for each single-step model. For both algorithms we focused on tuning the component that directly interacts with the single-step model: policy in MCTS and cost function in Retro*. Notably, we did not vary many of the other components of the algorithms (e.g. reward function in MCTS, value function in Retro*, search graph depth limits) to avoid an infeasibly large search space.

All tuning runs used 25 targets from the ChemBL Hard set used in~\citet{tripp2022reevaluating} and searched under a time limit of 5 minutes. As the primary objective we used the solve rate (i.e. number of solved targets), breaking ties first by the median and then mean number of non-overlapping routes found (formally, these three objectives were combined with weights $1.0$, $0.1$ and $0.01$, respectively). For each search algorithm and single-step model combination we ran $50$ trials using the default tuning algorithm in \texttt{optuna}~\citep{akiba2019optuna} to maximize the combined score.

For MCTS, we tuned the clipping range for the single-step model probabilities (lower bound in $[10^{-11}, 10^{-10}, ..., 10^{-5}]$, upper bound in $[0.9999, 0.999, 0.99, 0.9]$), temperature applied to the clipped distribution (in $[0.125, 0.25, ..., 4.0, 8.0]$), bound constant (in $[1, 10, 100, 1000, 10000]$) and node value constant (in $[0.25, 0.5, 0.75]$). For Retro*, we only tuned the clipping range (over the same values as for MCTS), as the temperature would have no effect due to using a constant-0 value function (referred to as Retro*-0 in \citet{chen20retrostar}).

In general, we found that the single-step probability clipping range has little effect on the algorithms, and so the performance of Retro* was not significantly improved through the hyperparameter tuning. Conversely, in MCTS parameters such as bound constant and temperature can have a sizable effect on the behaviour, and indeed choosing them carefully improved performance with respect to an initial guess. While MCTS seemingly performed worse than Retro* when using untuned hyperparameters, carefully setting the parameters of the former led it to perform on par with Retro*, echoing the conclusions from~\citet{tripp2022reevaluating}.

\section{Maintenance Plan for \textsc{Syntheseus}} \label{appendix:maintenance}

We intend to actively continue and support the development of \textsc{syntheseus} going forward,
including adding new features, fixing any bugs, and improving documentation.
As \textsc{syntheseus} is an open-source project on GitHub, anybody is free to modify and propose changes by raising an issue or opening a pull request.
We are committed to promptly responding to and engaging with all issues and pull requests.

The code to reproduce all experimental results (apart from those utilizing the proprietary Pistachio dataset) is publicly available.

\end{document}